\documentclass[runningheads]{llncs}
\usepackage{graphicx}
\usepackage{comment} 
\usepackage{amsmath,amssymb} % define this before the line numbering.
\usepackage{color}

\usepackage{caption}
\usepackage{subcaption}
\usepackage[x11names, svgnames, rgb]{xcolor}
\usepackage[utf8]{inputenc}
\usepackage{tikz}
\usetikzlibrary{arrows,shapes,positioning,arrows.meta,fit,calc,decorations.pathmorphing}
\usepackage{pgfplots}

\usepackage{multirow}
\usepackage{booktabs}
\usepackage{stackengine} 
\usepackage{dblfloatfix}
\usepackage{pifont}
\usepackage{mathtools}
\usepackage{wrapfig}

\usepackage{xspace}
\makeatletter
\DeclareRobustCommand\onedot{\futurelet\@let@token\@onedot}
\def\@onedot{\ifx\@let@token.\else.\null\fi\xspace}

\def\etal{\emph{et al}\onedot}
\@namedef{ver@everyshi.sty}{}
\makeatother

\newcommand\oast{\stackMath\mathbin{\stackinset{c}{0ex}{c}{0ex}{\ast}{\bigcirc}}}

\newcommand{\mytensor}[1]{\ensuremath{\mathcal{#1}}}
\newcommand{\myvector}[1]{\ensuremath{\mathbf{#1}}}

\begin{document}
% \linenumbers
\pagestyle{headings}
\mainmatter
\def\ECCVSubNumber{100}  % Insert your submission number here

\title{BATS: Binary ArchitecTure Search} % Replace with your title

\begin{comment}
\titlerunning{ECCV-20 submission ID \ECCVSubNumber} 
\authorrunning{ECCV-20 submission ID \ECCVSubNumber} 
\author{Anonymous ECCV submission}
\institute{Paper ID \ECCVSubNumber}
\end{comment}
%******************

%\begin{comment}
\titlerunning{BATS: Binary ArchitecTure Search}
\author{Adrian Bulat\inst{1}\orcidID{0000-0002-3185-4979} \and
Brais Martinez\inst{1}\orcidID{0000-0001-7511-8941} \and
Georgios Tzimiropoulos\inst{1,2}\orcidID{0000-0002-1803-5338}}
\authorrunning{Bulat et al.}
\institute{Samsung AI Center, Cambridge, UK \\
\email{adrian@adrianbulat.com, brais.mart@gmail.com}\\ \and
Queen Mary University of London, London, UK\\
\email{g.tzimiropoulos@qmul.ac.uk}}
%\end{comment}
%******************
\maketitle

\begin{abstract}
This paper proposes Binary ArchitecTure Search (BATS), a framework that drastically reduces the accuracy gap between binary neural networks and their real-valued counterparts by means of Neural Architecture Search (NAS). We show that directly applying NAS to the binary domain provides very poor results. To alleviate this, we describe, to our knowledge, for the first time, the 3 key ingredients for successfully applying NAS to the binary domain. Specifically, we (1) introduce and design a novel binary-oriented search space, (2) propose a new mechanism for controlling and stabilising the resulting searched topologies, (3) propose and validate a series of new search strategies for binary networks that lead to faster convergence and lower search times.
Experimental results demonstrate the effectiveness of the proposed approach and the necessity of searching in the binary space directly. Moreover, (4) we set a new state-of-the-art for binary neural networks on CIFAR10, CIFAR100 and ImageNet datasets.  Code will be made available.
\end{abstract}

\section{Introduction}\label{sec:introduction}

Network quantization and Network Architecture Search (NAS) have emerged as two important research directions with the goal of designing efficient networks capable of running on mobile devices. Network quantization reduces the size and computational footprint of the models by representing the activations and the weights using $N<32$ bits. Of particular interest is the extreme case of quantization, binarization, in which the model and the activations are quantized to a single bit~\cite{rastegari2016xnor,courbariaux2016binarized,courbariaux2015binaryconnect}. This allows to replace all floating point multiply-add operations inside a convolutional layer with bit-wise operations resulting in a speed-up of up to $57\times$~\cite{rastegari2016xnor}. Recent years have seen a progressive reduction of the performance gap with real-valued networks. However, research has almost exclusively focused on the ResNet architecture, rather than on efficient ones such as MobileNet. This is widely credited to pointwise convolutions not being amenable to binarization~\cite{bulat2017binarized,bethge2019back}.

As an orthogonal direction, NAS attempts to improve the overall performance by automatically searching for optimal network topologies using a various of approaches including evolutionary algorithms~\cite{xie2017genetic,real2019regularized}, reinforcement learning~\cite{zhong2018practical,zoph2018learning} or more recently by taking a differentiable view of the search process via gradient-based approaches~\cite{liu2018darts,liu2018progressive}. Such methods were shown to perform better than carefully hand-crafted architectures on both classification~\cite{liu2018progressive,liu2018darts,pham2018efficient,zoph2018learning} and fine-grained tasks~\cite{liu2019auto}. However, all of the aforementioned methods are tailored towards searching architectures for real-valued neural networks.

The aim of this work is to propose, for the first time, ways of reducing the accuracy gap between binary and real-valued networks via the state-of-the-art framework of differentiable NAS (DARTS)~\cite{liu2018darts,liu2018progressive}. 

We show that direct binarization of existing cells searched in the real domain leads to sub-par results due to the particularities of the binarization process, in which certain network characteristics typically used in real-valued models (e.g. $1\times1$ convolutions) may be undesired~\cite{bulat2017binarized,bethge2019back}. Moreover, we show that performing the search in the binary domain comes with a series of challenges of its own: it inherits and amplifies one of the major DARTS~\cite{liu2018darts} drawbacks, that of ``cell collapsing'', where the resulting architectures, especially when trained for longer, can result in degenerate cells in which most of the connections are skip connections~\cite{liu2018progressive} or parameter-free operations. In this work, we propose a novel Binary Architecture Search (BATS) method that successfully tackles the above-mentioned issues and sets a new state-of-the-art for binary neural networks on the selected datasets, yet simultaneously reducing computational resources by a wide margin. In summary, \textbf{our contributions} are:
\begin{enumerate}
   \item We show that directly applying NAS to the binary domain provides very poor results. To alleviate this, we describe, to our knowledge, for the first time, the 3 key ingredients for successfully applying NAS to the binary domain.
   \item We devise a novel search space specially tailored to the binary case and incorporate it within the DARTS framework (Section~\ref{ssec:method-binary-search-space}).
   \item We propose a temperature-based mechanism for controlling and stabilising the topology search (Section~\ref{ssec:method-regularization}).
   \item We propose and validate a series of new search strategies for binary networks that lead to faster convergence and lower search times (Section~\ref{ssec:method-strategy}).
   \item We show that our method consistently produces superior architectures for binarized networks within a lower computational budget, setting a new state-of-the-art on CIFAR10/100 and ImageNet datasets (Section~\ref{sec:experiments}).
\end{enumerate}
%%%\vspace*{-10pt}
\section{Related work}\label{sec:related-work}

\noindent\textbf{NAS:} While hand-crafted architectures significantly pushed the state-of-the-art in Deep Learning~\cite{howard2017mobilenets,sandler2018mobilenetv2,zhang2018shufflenet,ma2018shufflenet,hu2017squeeze,he2015deep}, recently, NAS was proposed as an automated alternative, shown to produce architectures that outperform the manually designed ones~\cite{real2017large}. NAS methods can be roughly classified in three categories: evolutionary-based~\cite{xie2017genetic,real2019regularized,real2017large}, reinforcement learning-based~\cite{baker2017designing,zhong2018practical,zoph2018learning} and, more recently, one-shot approaches, including differentiable ones~\cite{brock2018smash,liu2018darts,liu2019auto,cai2018proxylessnas}. The former two categories require significant computational resources during the search phase (3150 GPU-days for AmoebaNet~\cite{real2019regularized} and 1800 GPU-days for NAS-Net~\cite{zoph2018learning}). Hence, of particular interest in our work is the differentiable search framework (DARTS) of~\cite{liu2018darts} which is efficient. Follow-up work further improves on it by progressively reducing the search space~\cite{chen2019progressive} or by constraining the architecture parameters to be one-hot~\cite{xie2018snas}. Our method builds upon the DARTS framework incorporating the progressive training of~\cite{chen2019progressive}. In contrast to all the aforementioned methods that search for optimal real-valued topologies, to our knowledge, we are the first to study NAS for binary networks.

\noindent\textbf{Network binarization} is the most extreme case of network quantization in which weights and activations are represented with 1 bit. It allows for up to $57\times$ faster convolutions~\cite{rastegari2016xnor}, a direct consequence of replacing all multiplications with bitwise operations~\cite{courbariaux2015binaryconnect}. Typically, binarization is achieved by taking the sign of the real-valued weights and features~\cite{courbariaux2015binaryconnect,courbariaux2016binarized}. However, such an approach leads to large drops in accuracy, especially noticeable on large scale datasets. To alleviate this, Rastegari~\etal~\cite{rastegari2016xnor} introduce analytically computed real-valued scaling factors that scale the input features and weights. This is further improved in~\cite{bulat2019xnor} that proposes to learn the factors via back-propagation. In Bi-Real Net, Lin~\etal~\cite{liu2018bi} advocate the use of double-skip connections and of real-valued downsample layers. Wang~\etal~\cite{wang2019learning} propose a reinforcement learning-based approach to learn inter-channel correlations to better preserve the sign of convolutions. Ding~\etal~\cite{ding2019regularizing} introduce a distribution loss to explicitly regularize the activations and improve the information flow thought the network. While most of these methods operate within the same computational budget, another direction of research attempts to bridge the gap between binary and their real-valued networks by increasing the network size. For example, ABC-Net~\cite{lin2017towards} proposes to use up to $M=N=5$ branches, equivalent with running $M\times N$ convolutional layers in parallel, while Zhu~\etal~\cite{zhu2019binary} use an ensemble of up to 6 binary models. These methods expand the size of the network while preserving the general architecture of the ResNet~\cite{he2015deep} or WideResNet~\cite{zagoruyko2016wide}. In contrast to all the aforementioned methods in which binary architectures are hand-crafted, we attempt to automatically discover novel binary architectures, without increasing the computational budget. Our discovered architectures set a new state-of-the-art on the most widely used datasets.

\noindent\textbf{Very recent work on binary NAS} done concurrently with our work include~\cite{shen2019searching} and~\cite{singh2020learning}. As opposed to our work,~\cite{shen2019searching} simple searches for the number of channels in each layer inside a ResNet. ~\cite{singh2020learning} uses a completely different search space and training strategy. We note that we outperform significantly both of them in terms of accuracy/efficiency. On ImageNet, using  $1.55\times10^8$ FLOPs, we obtain a Top-1 accuracy of \textbf{66.1\%} vs 58.76\% using $1.63\times10^8$ FLOPS in~\cite{singh2020learning}. On CIFAR-10 we score a top-1 accuracy of 96.1\% vs 94.43\% in~\cite{shen2019searching}. While~\cite{shen2019searching} achieves an accuracy of 69.65\% on ImageNet, they use $6.6\times10^8$ FLOPS ($4.2\times$ more than our biggest model).

\section{Background}\label{sec:preliminaries}

\noindent\textbf{Network binarization:} We binarize the models using the method proposed by Rastegari~\etal~\cite{rastegari2016xnor} with the modifications introduced in~\cite{bulat2019xnor}: we learn the weight scaling factor $\alpha$ via back-propagation instead of computing it analytically while dropping the input feature scaling factor $\beta$. Assume a given convolutional layer $L$ with weights $\mytensor{W}\in \mathbb{R}^{o\times c\times w \times h}$ and input features $\mytensor{I}\in \mathbb{R}^{c\times w_{in} \times h_{in}}$, where $o$ and $c$ represent the number of output and input channels, ($w,h$) the width and height of the convolutional kernel, ($w_{in}\geq w$, $h_{in}\geq h$) the spatial dimensions of $\mytensor{I}$. The binarization is accomplished by taking the sign of $\mytensor{W}$ and $\mytensor{I}$ and then multiplying the output of the convolution by the trainable scaling factor $\alpha\in\mathbb{R}^+$:
\begin{equation}\label{eq:baseline}
		\mytensor{I} \ast \mytensor{W} \approx \left(\text{sign}(\mytensor{I}) \oast \text{sign}(\mytensor{W}) \right) \odot \myvector{\alpha},
\end{equation}
where $\odot$ denotes the element-wise multiplication, $\ast$ the real-valued convolution and $\oast$ its binary counterpart. During training, the gradients are used to update the real-valued weights $\mytensor{W}$ while the forward pass is done using $\text{sign}(\mytensor{W})$. Finally, we used the standard arrangement for the binary convolutional layer operations: Batch Norm, Sign, Convolution, Activation.

\noindent\textbf{DARTS:} Herein, we review DARTS~\cite{liu2018darts} and P-DARTS~\cite{chen2019progressive} upon which we build our framework: instead of searching for the whole network architecture, DARTS breaks down the network into a series of $L$ identical cells. Each cell can be seen as a Direct Acyclic Graph (DAG) with $N$ nodes, each representing a feature tensor. Within a cell, the goal is to choose, during the search phase, an operation $o(\cdot)$ from the predefined search space $O$ that will connect a pair of nodes. Given a pair of such nodes $(i,j)$ the information flow from $i$ to $j$ is defined as: $f_{i,j}(\mathbf{x}_i)=\sum_{o\in O}\frac{\text{exp}(\alpha_{i,j}^o)}{\sum_{o^{'}\in O} \text{exp}(\alpha_{i,j}^{o^{'}})}\cdot o(\mathbf{x}_i)$, where $\mathbf{x}_i$ is the output of the $i$-th node and $\alpha_{i,j}^o$ is an architecture parameter used to weight the operation $o(\mathbf{x}_i)$. The output of the node is computed as the sum of all inputs $\mathbf{x}_j=\sum_{i<j}f_{i,j}(\mathbf{x}_i)$ while the output of the cell is a depth-wise concatenation of all the intermediate nodes minus the input ones ($2,3,\dots,N-1$). During the search phase, the networks weights $\mytensor{W}$ and the architectures parameters $\alpha$ are learned using bi-level optimization where the weights and $\alpha$ are optimized on two different splits of the dataset. Because the search is typically done on a much shallower network due to the high computational cost caused by the large search space $O$, DARTS may produce cells that are under-performing when tested using deeper networks. To alleviate this, in~\cite{chen2019progressive} the search is done using a series of stages during which the worse performing operations are dropped and the network depth increases. For a complete detailed explanation see~\cite{liu2018darts} and~\cite{chen2019progressive}.

\section{Method}\label{sec:method}
This section describes the 3 key components proposed in this work: the binary search space, the temperature regularization and search strategy. As we also show experimentally in Section~\ref{ssec:ablation-effect}, we found that  \textit{all 3 components are necessary for successfully applying NAS to the binary domain}\footnote{When any of the components was not used, the obtained results were very poor.}.
%%%\vspace*{-5pt}
\subsection{Binary Neural Architecture Search Space}\label{ssec:method-binary-search-space}

Since searching across a large space $O$ is computationally prohibitive, most of the recent NAS methods manually define a series of 8 operations known to produce satisfactory results when used in hand-crafted network topologies: $3\times 3$ and $5\times 5$ dilated convolutions, $3\times 3$ and $5\times 5$ separable convolutions, $3\times 3$ max pooling, $3\times 3$ average pooling, identity (skip) connections plus the zero (no) connection. Table \ref{tab:search-space} summarizes the operations used for the real-valued case. Having an appropriate search space containing only desirable operations is absolutely essential for obtaining good network architectures: for example, a random search performed on the aforementioned space on CIFAR-10 achieves already 3.29\% top-error vs. 2.76\% when searched using second-order DARTS~\cite{liu2018darts}. 

\begin{table}[ht]
\caption{Comparison between the commonly used searched space for real valued networks and the proposed one, specially tailed for binary models.}
\label{tab:search-space}
\centering
\begin{tabular}{c|c}
\toprule
Real-valued & Ours (proposed)  \\
\midrule
Separable conv. ($3\times3)$ & Group convolution ($3\times3)$ \\
Separable conv. ($5\times5)$ & Group convolution ($5\times5)$ \\
\midrule
Dilated conv. ($3\times3)$  & Dilated Group conv. ($3\times3)$ \\
Dilated conv. ($5\times5)$ & Dilated Group conv.($5\times5)$ \\
\midrule
\multicolumn{2}{c}{Identity (\textit{skip connection)}} \\
\midrule
\multicolumn{2}{c}{Max pooling ($3\times 3$)} \\
\midrule
\multicolumn{2}{c}{Average pooling ($3\times 3$)} \\
\midrule
\multicolumn{2}{c}{Zero-op} \\
\bottomrule
\end{tabular}
%%%\vspace*{-15pt}
\end{table}
 
However, not all operations used for the real-valued case are suitable for searching binary architectures. In fact, we found that when using the standard DARTS search space, searching in the binary domain \textit{does not converge}. 
This is due to several reasons: the depth-wise convolutions are notoriously hard to binarize due to what we call the ``double approximation problem'': the real-valued depth-wise convolution is a ``compressed'' approximate version of the normal convolution and, in turn, the binary depth-wise in a quantized approximation of the real-valued one. Furthermore, the $1\times 1$ convolutional layers and the bottleneck block~\cite{he2015deep} were already shown to be hard to binarize~\cite{bulat2017binarized} because the features compression that happens in such modules amplifies the high information degradation already caused by binarization. Both $1\times 1$ and separable convolutions are present in the current search space typically used for search real-valued architectures (see Table~\ref{tab:search-space}), making them inappropriate for binarization. Moreover, while the dilated convolution contains 2 serialized convolutions, the separable one contains 4 which (a) causes discrepancies in their convergence speed and (b) can amplify the gradient fading phenomena that often happens during binarization.

To this end, we propose a new search space $O$, shown in Table \ref{tab:search-space}, constructed from a binary-first point of view that avoids or alleviates the aforementioned shortcomings. While we preserve the zeros and identity connections alongside the $3\times 3$ max and average pooling layers which do not contain learnable parameters or binary operations, we propose to replace all the convolutional operations with the following new ones: $3\times 3$ and $5\times 5$ grouped convolutions and dilated grouped convolutions, also with a kernel size of $3\times 3$ and $5\times 5$. This removes all the $1\times 1$ convolutional layers directly present in the cell's search space while maintaining the efficiency via the usage of grouped convolutions. 

We note that there is a clear trade-off here between efficiency and accuracy controlled via the group size and number of channels. In this work, and in contrast to other works that rely on grouped convolutions (e.g. \cite{zhang2018shufflenet}), we propose to use a very high group size which leads to behaviours closer to that of a depth-wise convolution (a small one will come at a price of higher computational budgets, while the extreme case \#groups=\#in\_channels will again exacerbate the double-approximation problem). We used a group size of 12 and 3 (only) channels per group (totalling 36 channels) for CIFAR, and a group size of 16 and 5 (only) channels per group (totalling 80 channels) for ImageNet. 
Note that we also explored the effect of adding a channel shuffle layer~\cite{zhang2018shufflenet} after each grouped convolution, typically used to enable cross-group information flow for a set of group convolution layers. However, due to the fact that \#groups $\gg$ \#channels per group, they were unable to offer accuracy gains. Instead, we found that the $1\times 1$ convolutions present between different DARTS cells are sufficient for combining information between the different groups.

Furthermore, in the proposed search space, in all convolutional layers, the depth of all operations used is equal to 1 meaning that we used 1 convolution operator per layer, as opposed to 4 convolutions (2 depth-wise separable) used in DARTS, which facilitates learning. Another consequence of using operations with a depth equal to 1 is a latency improvement due to effectively a shallower network. For a visual illustration of differences between DARTS and the proposed BATS please check the supplementary material.

Finally, to improve the gradient and information flow, which is even more critical for the case of binary networks, while also speeding-up the convergence of such operations during the  search phase, we explicitly add an identity connection on each convolutional operation such that $f_{i,j}(\mathbf{x}_i)=f_{i,j}(\mathbf{x}_i)+\mathbf{x}_i$. 

The efficacy of the proposed search space is confirmed experimentally in Table~\ref{tab:ev_cifar}. As the results from Table~\ref{tab:ev_cifar} show, the architectures found by searching within the proposed search space constantly outperform the others.  

%%%\vspace*{-5pt}
\subsection{Search Regularization and Stabilisation}\label{ssec:method-regularization}

Despite its success and appealing formulation, DARTS accuracy can vary wildly between runs depending on the random seed. In fact, there are cases in which the architectures found  perform worse than the ones obtained via random search. Furthermore, especially when trained longer or if the search is performed on larger datasets DARTS can converge towards degenerate solutions dominated by skip connections~\cite{chen2019progressive}. While in~\cite{chen2019progressive} the authors propose to address this by a) applying a dropout on the skip connections during the architecture search and b) by preserving a maximum of 2 skip connections per cell as a post processing step that simply promotes the operation with the second highest probability, we found that such mechanism can still result in a large amount of randomness and is not always effective: for example, it can replace skip connections with pooling layers (which have no learning capacity) or the discovered architecture might even already contain too few skip connections.

\begin{wrapfigure}[19]{r}{0.5\textwidth}
	\centering
	%%%\vspace{-0.3cm}
  %\scalebox{0.75}{\input{figures/arch_params/reduce.tex}}
  \scalebox{0.73}{\includegraphics{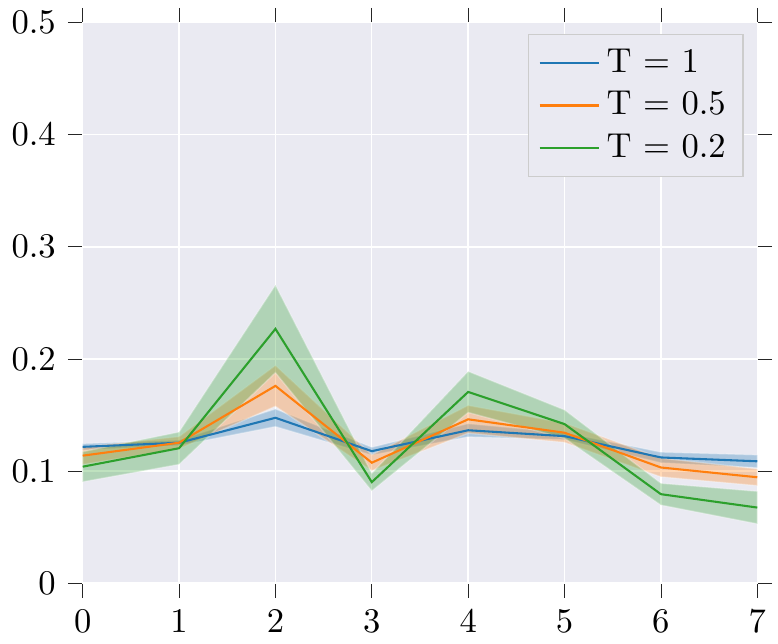}}
	\caption{Distribution of architecture parameters after the first stage for a given cell (data points were sorted by magnitude and do not correspond to the same ops): for low temperatures the network is forced to be more discriminative.}
	\label{fig:reduce-cell-arch-distribution}
\end{wrapfigure}

 Such problems are even more noticeable when the search is performed in the binary domain directly. Given that, during search, the input to the node $j$ is obtained by taking a weighted sum of all incoming edges, in order to maximise the flow of information, the architecture parameters $\alpha$ tend to converge to the same value making the selection of the final architecture problematic and susceptible to noise resulting in topologies than may perform worse than a random selection. Furthermore, the search is highly biased towards real-valued operations (pooling and skip connections) that at early stages can offer larger benefits. 

To alleviate the aforementioned issues and encourage the search procedure to be more discriminative forcing it to make ''harder`` decisions, we propose to use a temperature factor $T<1$ defining the flow from node $i$ to $j$ as follows:
\begin{equation}
   f_{i,j}(\mathbf{x}_i)=\sum_{o\in O}\frac{\text{exp}(\alpha_{i,j}^o/T)}{\sum_{o^{'}\in O} \text{exp}(\alpha_{i,j}^{o^{'}}/T)}\cdot o(\mathbf{x}_i).
\end{equation}

This has the desirable effect of making the distribution of the architecture parameters less uniform and spikier (i.e. more discriminative). Hence, during search, because the information streams are aggregated using a weighted sum, the network cannot equally (or near-equally) rely on all possible operations by pulling information from all of them. Instead, in order to ensure convergence to a satisfactory solution it has to assign the highest probability to a non 0-ops path, enforced by a sub-unitary temperature ($T<1$). This behaviour also follows closer the evaluation procedure where a single operation will be selected, reducing as such the performance discrepancy between the search, where the network pulls information from all paths, and evaluation.

\begin{wrapfigure}[23]{r}{0.5\textwidth}
	\centering
	%%%\vspace{-0.3cm}
  %\scalebox{0.75}{\input{figures/arch_params/normal_arch_ops.tex}} 
  \scalebox{0.75}{\includegraphics{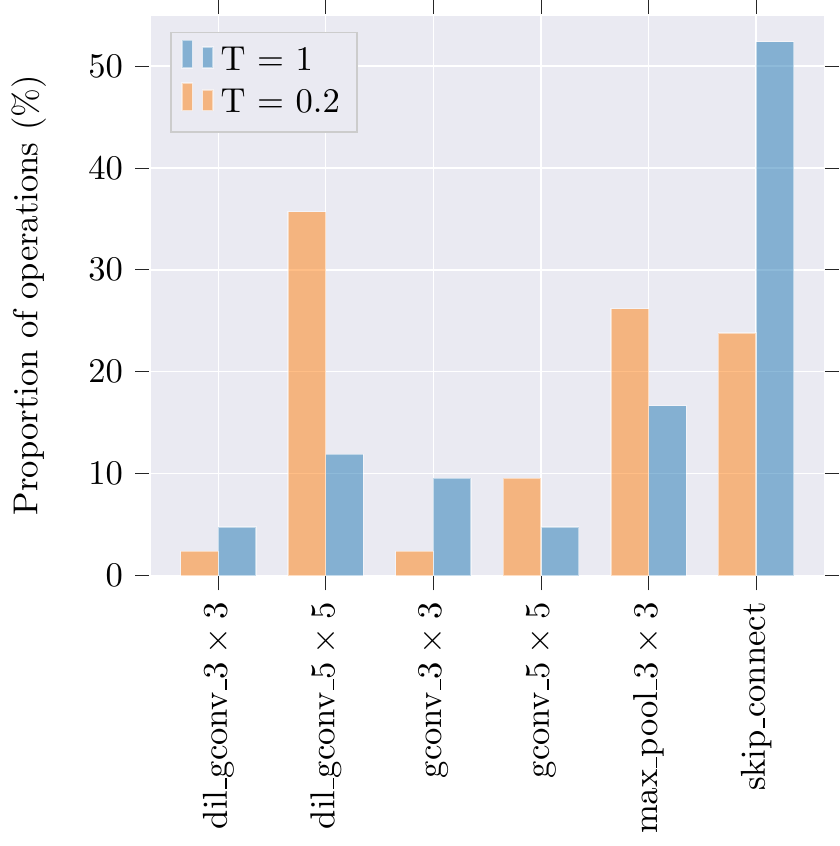}}
	\caption{Probability of a given op. in a cell at the end of the search process for different temperatures: for $T=0.2$ there is a significant increase in the number of $5\times 5$ convolutions and a decrease in the number of skip connections. }
	\label{fig:normal-cell-ops-distribution}
\end{wrapfigure}

Fig.~\ref{fig:reduce-cell-arch-distribution} depicts the distribution of the architecture parameters for a given cell for different temperatures. For low temperatures, the network is forced to make more discriminative decisions which in turn makes it rely less on identity connections. This is further confirmed by Fig.~\ref{fig:normal-cell-ops-distribution} which depicts the  chance of encountering a given op. in a normal cell at the end of the search process for different temperatures. 
%%%\vspace*{-5pt}
\subsection{Binary Search Strategy}\label{ssec:method-strategy}

Despite their appealing speed-up and space saving, binary networks remain harder to train compared to their real-valued counterparts, with methods typically requiring a pre-training stage~\cite{liu2018bi} or carefully tuning the hyper-parameters and optimizers~\cite{rastegari2016xnor,courbariaux2015binaryconnect}. For the case of searching binary networks, directly attempting to effectuate an architecture search using binary weights and activations in most of our attempts resulted either in degenerate topologies or the training simply converges to extremely low accuracy values. We also note that, as expected and as our experiments have confirmed, performing the search in the real domain directly and then binarizing the network is sub-optimal.

To alleviate the problem, we propose a two-stage optimization process in which during the search the activations are binarized while the weights are kept real, and once the optimal architecture is discovered we proceed with the binarization of the weights too during the evaluation stage\footnote{More specifically, during evaluation, we first train a new network with binary activations and real-valued weights from scratch, and then binarize the weights. Then, the fully binarized network is evaluated on the test set.}. This is motivated by the fact that while the weights of a real-valued network can be typically binarized without a significant drop in accuracy, the same cannot be said about the binarization of the activations where due to the limited number of possible states the information flow inside the network drops dramatically. Hence, we propose to effectively split the problem into two sub-problems: weight and feature binarization and during the search we try to solve the hardest one, that is the binarization of the activations. Once this is accomplished, the binarization of the weights following always results in little drop in accuracy ($\sim1\%$).

%%%\vspace*{-15pt}
\begin{figure*}
     \centering
     \begin{subfigure}[b]{0.45\textwidth}
         \centering
        % \begin{tikzpicture}[>=latex',line join=bevel]
        %     \useasboundingbox (0,0) rectangle (7.5,3);
        %     \scope[transform canvas={scale=.25}]
        %     \pgfsetlinewidth{1bp}
        %     \input{figures/cells/ours_normal_binary.tex}
        %     \endscope
        % \end{tikzpicture}
        \includegraphics{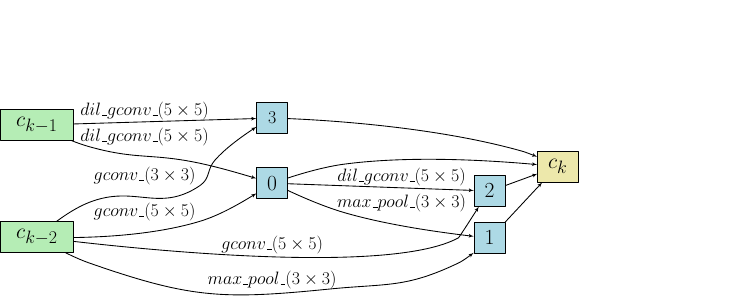}
        \caption{Normal binary cell}
        \label{fig:ours_normal_cell}
     \end{subfigure}
     \hfill
     \begin{subfigure}[b]{0.45\textwidth}
        \centering
        % \begin{tikzpicture}[>=latex',line join=bevel]
        %     \useasboundingbox (0,0) rectangle (7.5,3);
        %     \scope[transform canvas={scale=.25}]
        %     \pgfsetlinewidth{1bp}
        %     \input{figures/cells/ours_reduce_binary.tex}
        %     \endscope
        % \end{tikzpicture}
        \includegraphics{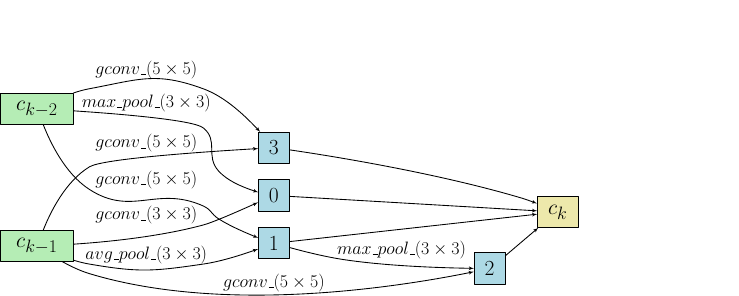}
        \caption{Reduction binary cell}
        \label{fig:ours_reduction_cell}
     \end{subfigure}
     \caption{Binary normal (a) and reduction (b) cells discovered by our method. Notice the prevalence of large kernels ($5 \times 5$) and of the wider reduction cell, both contributing to increasing the filters' diversity and the flow of information.}
     \label{fig:cells}
\end{figure*}
%%%\vspace*{-15pt}
\section{Ablation studies}\label{sec:ablation}
%%%\vspace*{-7pt}
\subsection{Effect of the proposed components}\label{ssec:ablation-effect}
Herein, we analyse the contribution to the overall performance of each proposed component: the newly introduced search space for binary networks (Section~\ref{ssec:method-binary-search-space}), the effect of the temperature adjustment (Section~\ref{ssec:method-regularization}) and finally, of performing the search using real-valued weights and binary activations (Section~\ref{ssec:method-strategy}). 
%\begin{table}[ht]
\begin{wraptable}[9]{r}{0.5\textwidth}
%%%\vspace{-0.2cm}
\caption{Effect of search space on Top-1 accuracy on CIFAR-10.}
%%%\vspace{-0.1cm}
\label{tab:search-ablation}
\centering
\begin{tabular}{lll}
\toprule
Search space & Temperature & Accuracy \\
\midrule
Ours & Ours  & $\mathbf{93.7\pm0.6}$ \\
\cite{liu2018darts} & Ours & $51.0\pm7.5$ \\
Ours & \cite{liu2018darts} & $\mathbf{86.0\pm2.0}$  \\
\bottomrule
\end{tabular}
\end{wraptable}
We emphasize that \textit{all 3 ingredients are needed in order to stabilize training and obtain good accuracy}. This suggests that it is not straightforward to show the effect of each component in isolation. To this end, we always \textit{keep fixed} 2 of the proposed improvements and vary the other component to understand its impact. We also note that in all cases the final evaluated networks are fully binary (i.e. both the weights and the activations are binarized).
\setlength{\tabcolsep}{4pt}
%\begin{table}[ht]
\begin{wraptable}[9]{r}{0.5\textwidth}
%%%\vspace{-0.1cm}
\caption{Effect of temperature on Top-1 accuracy on CIFAR-10. T=1 was used in DARTS~\cite{liu2018darts}.  *Often leads to degenerate solution (i.e. all skip connections).}
\label{tab:temperature-effect}
\centering
\begin{tabular}{lcccc}
\toprule
Temperature &  1 & 0.5 & 0.2 & 0.05 \\
\midrule
Accuracy & 86.0* & 92.1 & 93.7  & 93.3 \\
\bottomrule
\end{tabular}
%\vspace{-3pt}
\end{wraptable}
\setlength{\tabcolsep}{1.4pt}

\noindent\textbf{Importance of new search space:} 
We keep the temperature from Section~\ref{ssec:method-regularization} and search strategy 
from Section~\ref{ssec:method-strategy} fixed and then compare the proposed search space with that used in~\cite{liu2018darts}. As the results from Table~\ref{tab:search-ablation} show, searching for binary networks using the search space of~\cite{liu2018darts} leads to much worse results ($\sim51\%$ vs.$\sim94\%$) and a high variation in the performance of the networks (a std. of 7.5\% across 5 runs). This clearly shows that the previously used search space is not suitable for binary networks.
\newline\textbf{Impact of temperature:} We keep the search space from Section~\ref{ssec:method-binary-search-space} and search strategy from Section~\ref{ssec:method-strategy} fixed and evaluate the use of the temperature proposed in Section~\ref{ssec:method-regularization}. As Table~\ref{tab:temperature-effect} shows, a decrease in temperature has a direct impact on discovering of higher performing models. At the opposite spectrum, a low temperature often leads to degenerate solutions. Note that decreasing the temperature further has a negative impact on the accuracy, because, as the temperature goes to 0, the distribution of the architecture parameters resembles a delta function, which hinders the training process. 
\begin{wrapfigure}[17]{r}{0.5\textwidth}
    %%%\vspace{-0.3cm}
	\centering
  %\scalebox{0.67}{\input{figures/results/cifar100_aug.tex}}
  \scalebox{0.67}{\includegraphics{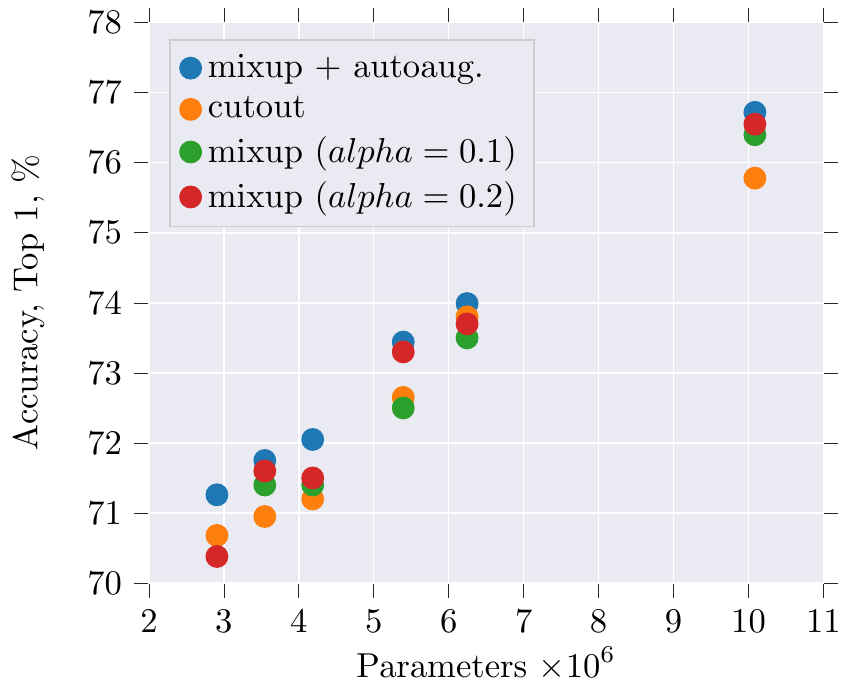}}
	\caption{Effect of data augmentation: Top-1 accuracy on CIFAR-100 of the network constructed using the discovered cell for different number of network parameters.}
	\label{fig:cifar100}
\end{wrapfigure}

\noindent\textbf{Importance of search strategy:} We keep the search space from Section~\ref{ssec:method-binary-search-space} and temperature from Section~\ref{ssec:method-regularization} fixed and evaluate the effect of the proposed search strategy from Section~\ref{ssec:method-strategy}. Table~\ref{tab:search-cost} summarizes the different strategies for searching binary architectures and their properties. The first row represents the case where the search is done in the real domain and then binarization follows. As the second row shows, if the search is done in the binary domain for both weights and activations, the search does not converge. As the third row shows, searching in the binary domain for the weights while keeping the activations real converges but results in low accuracy networks. Finally, the last row shows the proposed search strategy where the search is done in the binary domain for the activations and in the real domain for the weights. This configuration is the only one that yields stable search procedure and high accuracy networks (see std). Note that for all cases, the architectures are fully binarized (i.e. both activations and weights) prior to evaluation on the test set.
\begin{table}[ht]
\caption{Different strategies for searching binary architectures and their properties. The last row represents the proposed strategy. \textbf{All resulting topologies were trained using binary weights and activations during evaluation.}}
\label{tab:search-cost}
\centering
\begin{tabular}{lcccc}
\toprule
\multicolumn{2}{c}{\textbf{Search domain}} & \textbf{Search cost} & \textbf{\multirow{2}{*}{Stable}} & \textbf{\multirow{2}{*}{Accuracy}} \\
\cmidrule(lr){0-1}
 W &  A  & (GPU-days) &  & \\
\midrule
Real & Real & 0.2 & Yes & $88.4\pm0.8$ \\
Binary & Binary & 0.3 & No & $84.6\pm1.7$ \\
Binary & Real & 0.25 & Yes & $91.1\pm1.7$ \\
\midrule
Real & Binary & 0.25 & Yes & $\mathbf{93.7\pm0.6}$\\
\bottomrule
\end{tabular}
\end{table}
%\noindent
%%%\vspace*{-7pt}
\subsection{Impact of augmentation} While the positive impact of data augmentation has been explored  for the case of real-valued networks, to the best of our knowledge, there is little to no work on this for the case of binary networks. Despite the fact that binarization is considered to be an extreme case of regularization~\cite{courbariaux2015binaryconnect}, we found that augmentation is equally important, and confirm that most of the augmentations applied to real-valued networks, can also improve their binarized counterparts. See Fig.~\ref{fig:cifar100}.
%%%\vspace{-3pt}
\section{Experiments}\label{sec:experiments}
We conducted experiments on three different image classification datasets, namely CIFAR10~\cite{krizhevsky2009learning}, CIFAR100~\cite{krizhevsky2009learning} and ImageNet~\cite{deng2009imagenet}. Note that the architecture search is carried out on CIFAR-10 and tested on all three. 

%%%\vspace{-5pt}
\subsection{Architecture search}\label{ssec:architecture-search}

\noindent \textbf{Implementation details:} The training process consists of 3 stages during which the network depth is increased from 5 to 11 and finally 17 cells. Concomitantly, at each stage, the search space is gradually reduced to 8, 5 and respectively 3 ops. As opposed to~\cite{chen2019progressive}, we did not use \textit{on-skip} dropout since we found that it leads to unstable solutions.  Instead, we used the mechanism proposed in Section~\ref{ssec:method-regularization}, setting the temperature for the normal and reduction cells to $0.2$ and $0.15$ respectively. The latter is lower to encourage wider cells. During each stage, the network is trained using a batch size of 96 for 25 epochs. For the first 10 epochs only the network weights are learned. The remaining 15 epochs update both the architecture and network parameters. All the parameters are optimized using Adam~\cite{kingma2014adam}. The learning rate was set to $\eta=0.0006$, weight decay to $w_d=0.001$ and momentum to $\beta= (0.5,0.999)$ for the architecture parameters and $\eta=0.001$, $w_d=0.0003$, $\beta= (0.9, 0.999)$ for the network parameters. To keep the search time low, we used the first-order optimization approach of DARTS. All methods were implemented in PyTorch~\cite{paszke2017automatic}.

\noindent\textbf{Discovered topologies:} Overall, our approach is capable of discovering high-performing binary cells, offering new or validating existing insights about optimal binary topologies. As depicted in Fig.~\ref{fig:cells}, the cells found tend to prefer convolutional layers with larger kernel sizes ($5\times 5$) which help alleviating the limited representational power found in such networks (a $3\times 3$ kernel has maximum number of $2^9$ unique binary filters while a $5\times 5$ has $2^{25}$). In addition, to preserve the information flow, a real-valued  path that connects one of the input nodes to  the output is always present.

Furthermore, since the down-sampling operation compresses the information across the spatial dimension, to compensate for this, the reduction cell tends to be wider (i.e. more information can flow through) as opposed to the normal cell which generally is deeper. As Table~\ref{tab:ev_cifar} shows, our searched architecture outperforms all prior ones searched in the real-valued domain and then binarized.
%%%\vspace{-10pt}
\begin{table}[ht]
%\begin{threeparttable}
\caption{Comparison with state-of-the-art NAS methods on CIFAR10 and CIFAR100. For all methods we apply the binarization process described in Section~\ref{sec:preliminaries} on the best cell provided by the authors.} %Notice that our approach significantly outperforms all prior methods.}
\label{tab:ev_cifar}
\centering
\begin{tabular}{lccccc}
\toprule
\textbf{\multirow{2}{*}{Architecture}} & \multicolumn{2}{c}{\textbf{Test Acc. (\%})} & \textbf{Params} & \textbf{Search Cost} & \textbf{\multirow{2}{*}{ \begin{tabular}{@{}c@{}}Search \\ Method\end{tabular}}} \\
\cmidrule(lr){2-3}
&                            \textbf{C10} & \textbf{C100} & \textbf{(M)} & \textbf{(GPU-days)} &\\
\midrule
ResNet-18~\cite{he2015deep}                       & 91.0 & 66.3 & 11.2 & -    & manual \\
Random search                      & 92.9 & 70.2 & - & -    & random \\
\midrule
NASNet-A~\cite{zoph2018learning}                 & 90.3 & 66.1     & 3.9  & 1800 & RL      \\
AmoebaNet-A~\cite{real2019regularized}           & 91.5 & 65.6     & 3.2  & 3150 & evolution \\
\midrule
DARTS (first order)~\cite{liu2018darts}           & 89.4 & 63.3 & 3.3 & 1.5  & grad-based \\
DARTS (second order)~\cite{liu2018darts}          & 91.1 & 64.0 & 3.3 & 4.0 & grad-based \\
P-DARTS~\cite{chen2019progressive}                                   & 89.5 & 63.9 & 3.6  & 0.3 & grad-based \\
\midrule
Ours                          & 93.7 & 70.7 & 2.8 & 0.25 & grad-based \\
Ours + autoaug.~\cite{cubuk2018autoaugment} & 94.1 & 71.3 & 2.8 & 0.25 & grad-based \\
Ours (medium)                          & 94.6 & 73.5 & 5.4 & 0.25 & grad-based \\
Ours (medium)  + autoaug.~\cite{cubuk2018autoaugment}                          & 94.9 & 74.0 & 5.4 & 0.25 & grad-based \\
Ours (large)                          & 95.5 & 75.7 & 10.0 & 0.25 & grad-based \\
Ours (large)  + autoaug.~\cite{cubuk2018autoaugment}                          & 96.1 & 76.8 & 10.0 & 0.25 & grad-based \\
\bottomrule
\end{tabular}
\end{table}
%%%\vspace{-17pt}
\subsection{Comparison against state-of-the-art}

In this section, we compare the performance of the searched architecture against state-of-the-art network binarization methods on CIFAR10/100 and ImageNet.

\noindent
\textbf{Implementation details:} This section refers to training the discovered architectures from scratch for evaluation purposes. For all models we use Adam with a starting learning rate of $0.001$, $\beta=(0.9,0.999)$ and no weight decay. The learning rate was reduced following a cosine scheduler~\cite{loshchilov2016sgdr}. In line with other works~\cite{liu2018darts}, on CIFAR10 and CIFA100, the models were trained for 600 epochs while on ImageNet for 75. The batch size was set to 256. The drop-path~\cite{larsson2016fractalnet} was set to 0 for all experiments since binary networks are less prone to overfitting and already tend to use all parallel paths equally well. Following~\cite{liu2018darts}, we add an auxiliary tower~\cite{szegedy2015going} with a weight of 0.4. Unless otherwise stated, the network trained consisted of 20 cells and 36 initial channels with a group size of 12 channels for CIFAR10/100 and respectively 80 initial channels and 14 cells for ImageNet. For data augmentation, similarly to~\cite{liu2018darts,chen2019progressive}, we padded the images appropriately on CIFAR10/100 and applied CutOut~\cite{devries2017improved} with a single hole and a length of 16px for CIFAR-10 and Mix-up ($\alpha=0.2)$ for CIFAR-100 respectively. For ImageNet, we simply resized the images to $256\times 256$px, and then randomly cropped them to $224\times 224$px. During testing, the images were center-cropped.
\newline\newline\textbf{CIFAR10:} When compared against network topologies discovered by existing state-of-the-art NAS approaches, as the results from Table~\ref{tab:ev_cifar} show, our method significantly outperforms all of them on both CIFAR-10 and CIFAR-100 datasets. When using 20\% the parameters than the competing methods, our approach offers gains of $5$ to $7$\% on CIFAR-100 and $2.5-4$\% on  CIFAR-10. This clearly confirms the importance of the proposed search space and strategy.
Furthermore, we test how the discovered architecture performs when scaled-up. In order to increase the capacity of the models we concomitantly adjust the width and \#groups of the cells. By doing so, we obtain a $2$\% improvement of CIFAR-10 and $5$\% on CIFAR-100 almost matching the performance of a real-valued ResNet-18 model. We note that for fairness all other topologies we compared against where modified using the structure described in Section~\ref{sec:preliminaries}.
\newline For an exhaustive comparison on CIFAR-10 see the supplementary material.
\newline\newline\textbf{ImageNet:} Herein, we compare our approach against related state-of-the-art binarization and quantization methods. As the results from Table~\ref{tab:sota-imagenet-all} show, our discovered architecture outperforms all existing binarization methods while using a lower computational budget for the normal thin model and offers a gain of 5\% for the 2x-wider one. Furthermore, our methods compared favorable even against methods that either significantly increase the network computational requirements or use more bits for quantization. 
%%%\vspace{-10pt}
\begin{table}[!ht]
\caption{Comparison with selected SOTA binary methods on ImageNet in terms of computational cost. Notice, that within a similar budget our method achieves the highest accuracy. For a full comparison see Table~\ref{tab:sota-imagenet-all}.}
\label{tab:sota-imagenet}
\centering
\setlength\tabcolsep{3.5pt}
    \begin{tabular}{lccccc}
    \toprule
    \textbf{\multirow{2}{*}{Architecture}} & \multicolumn{2}{c}{\textbf{Accuracy (\%})} & \multicolumn{2}{c}{\textbf{\# Operations}} & \textbf{\multirow{2}{*}{\# bits}} \\
    \cmidrule(lr){2-3}
    \cmidrule(lr){4-5}
    &                            \textbf{Top-1} & \textbf{Top-5} & \textbf{FLOPS$\times10^8$} & \textbf{BOPS$\times10^9$} & (W/A)\\
    \midrule
    BNN~\cite{courbariaux2016binarized} & 42.2 & 69.2 & $1.314$ & $1.695$ & 1/1 \\
    XNOR-Net~\cite{rastegari2016xnor} & 51.2 & 73.2 & $1.333$ & $1.695$ & 1/1 \\
    CCNN~\cite{xu2019accurate} & 54.2 & 77.9 & $1.333$ & $1.695$ & 1/1 \\
    Bi-Real Net~\cite{liu2018bi} & 56.4 & 79.5 & $1.544$ & $1.676$ & 1/1 \\
    XNOR-Net++~\cite{bulat2019xnor} & 57.1 & 79.9 & $1.333$ & $1.695$ & 1/1 \\
    CI-Net~\cite{wang2019learning} & 59.9 & 84.2 & $-$ & $-$ & 1/1 \\
    \midrule
    \textbf{BATS (Ours)}                         & 60.4 & 83.0 & $0.805$ & $1.149$& 1/1 \\
    \textbf{BATS [2x-wider] (Ours)}                         & \textbf{66.1} & \textbf{87.0} & $1.210$ & $2.157$ & 1/1 \\
    \bottomrule
    \end{tabular}
\end{table}
%%%\vspace{-15pt}
\subsection{Network efficiency}\label{ssec:network-efficiency}

The current most popular settings of binarizing neural networks preserve the first and last layer real-valued. However, for the currently most popular binarized architecture, i.e. ResNet-18~\cite{he2015deep}, the first convolutional layer accounts for approx. 6.5\% of the total computational budget ($1.2\times10^8$ FLOPs out of the total of $1.8\times 10^9$ FLOPs), a direct consequence of the large kernel size ($7\times 7$) and the high input resolution. In order to alleviate this, beyond using the ImageNet stem cell used in DARTS~\cite{liu2018darts} that replaces the $7\times 7$ layer with two $3\times 3$, we replaced the second convolution with a grouped convolution ($g=4$).  This alone allows us to more than halve the number of real-valued operations (see Table~\ref{tab:sota-imagenet}). Additionally, in order to transparently show the computational cost of the tested models, we separate the binary operations (BOPs) and FLOPs into two distinctive categories. Furthermore, as Table~\ref{tab:sota-imagenet} shows, our method surpasses sophisticated approaches while having a significantly lower number of ops.
%%%\vspace{-10pt}
\begin{table}[!ht]
\caption{Comparison with state-of-the-art binarization methods on ImageNet, including against approaches that use low-bit quantization (upper section) and ones that increase the capacity of the network (middle section).  For the latter case the last column also indicates the capacity scaling used. Models marked with ** use real-valued downsampling convolutional layers.}
\label{tab:sota-imagenet-all}
\centering
\setlength\tabcolsep{3.5pt}
    \begin{tabular}{lccc}
    \toprule
    \textbf{\multirow{2}{*}{Architecture}} & \multicolumn{2}{c}{\textbf{Accuracy (\%})} & \textbf{\multirow{2}{*}{\# bits}} \\
    \cmidrule(lr){2-3}
    &                            \textbf{Top-1} & \textbf{Top-5}  & (W/A)\\
    \midrule
    BWN~\cite{courbariaux2016binarized} & 60.8 & 83.0  & 1/32 \\
    TTQ~\cite{zhu2016trained} & 66.6 & 87.2  & 2/32 \\
    HWGQ~\cite{cai2017deep} & 59.6 & 82.2  & 1/2 \\
    LQ-Net~\cite{zhang2018lq} & 59.6 & 82.2  & 1/2 \\
    SYQ~\cite{faraone2018syq} & 55.4 & 78.6  & 1/2 \\
    DOREFA-Net~\cite{zhou2016dorefa} & 62.6 & 84.4  & 1/2 \\
    \midrule
    ABC-Net ($M,N=1$)~\cite{lin2017towards} & 42.2 & 67.6 & 1/1 \\
    ABC-Net ($M,N=5$)~\cite{lin2017towards} & 65.0 & 85.9  & (1/1)$\times 5^2$ \\
    Struct. Approx.~\cite{zhuang2019structured} & 64.2 & 85.6  & (1/1)$\times 4$ \\
    Struct. Approx.**~\cite{zhuang2019structured} & 66.3 & 86.6  & (1/1)$\times 4$ \\
    CBCN~\cite{liu2019circulant} & 61.4 & 82.8  & (1/1)$\times 4$ \\
    Ensamble~\cite{zhu2019binary} & 61.0 & -  & (1/1)$\times 6$ \\
    \midrule
    BNN~\cite{courbariaux2016binarized} & 42.2 & 69.2  & 1/1 \\
    XNOR-Net~\cite{rastegari2016xnor} & 51.2 & 73.2  & 1/1 \\
    CCNN~\cite{xu2019accurate} & 54.2 & 77.9  & 1/1 \\
    Bi-Real Net**~\cite{liu2018bi} & 56.4 & 79.5  & 1/1 \\
    Rethink. BNN**~\cite{helwegen2019latent} & 56.6 & 79.4  & 1/1 \\
    XNOR-Net++~\cite{bulat2019xnor} & 57.1 & 79.9  & 1/1 \\
    CI-Net**~\cite{wang2019learning} & 59.9 & 84.2 & 1/1 \\
    \textbf{BATS (Ours)}                         & 60.4 & 83.0 &  1/1 \\
    \textbf{BATS [2x-wider] (Ours)}                         & \textbf{66.1} & \textbf{87.0} & 1/1 \\
    \bottomrule
    \end{tabular}
\end{table}
%%%\vspace{-10pt}
\section{Conclusion}
 In this work we introduce a novel Binary Architecture Search (BATS) that drastically reduces the accuracy gap between binary models and their real-valued counterparts, by searching for the first time directly for binary architectures. To this end we, (a) designed a novel search space specially tailored to binary networks, (b) proposed a new regularization method that helps stabilizing the search process, and (c) introduced an adapted search strategy that speed-ups the overall network search. Experimental results conducted on CIFAR10, CIFAR100 and ImageNet demonstrate the effectiveness of the proposed approach and the need of doing the search in the binary space directly. 

%\clearpage
\bibliographystyle{splncs04}
\bibliography{egbib}

\appendix
\section{Additional comparison with state-of-the-art on CIFAR10}

To further showcase the improvements offered by our approach, herein we compare its performance against an additional set of state-of-the-art methods on the CIFAR-10 dataset. As the results from Table~\ref{tab:sota-cifar10-all} show, our method significantly outperforms all previous ones across different architectures(VGG, ResNet, WRN) and quantization levels.

\begin{table}[!htpb]
%\begin{threeparttable}
\begin{center}
%\resizebox{\columnwidth}{!}{%
\setlength\tabcolsep{3.5pt}
    \begin{tabular}{lccc}
    \hline
    \textbf{Method} & \textbf{Acc.(\%}) & Architecture & \# bits (W/A) \\
    \hline
    BC~\cite{courbariaux2015binaryconnect} & 90.1 & VGG-small & 1/32 \\
    TTQ~\cite{zhu2016trained} & 91.1 & ResNet-20 & 2/32 \\
    HWGQ~\cite{cai2017deep} & 92.5 & VGG-small & 1/2 \\
    LQ-Net~\cite{zhang2018lq} & 93.4 & VGG-small & 1/2 \\
    \hline
    CBCN~\cite{liu2019circulant} & 91.6 & ResNet-18  & (1/1)$\times 4$ \\
    CBCN~\cite{liu2019circulant} & 93.4 & WRN40  & (1/1)$\times 4$ \\
    BNN~\cite{courbariaux2016binarized} & 89.9 & VGG-small & 1/1 \\
    XNOR-Net~\cite{rastegari2016xnor} & 89.8 & VGG-small & 1/1 \\
    CCNN~\cite{xu2019accurate} & 92.3 & VGG-small  & 1/1 \\
    CI-Net~\cite{wang2019learning} & 92.5 & VGG-small & 1/1 \\
    \hline
    \textbf{BATS (Ours)}                         & \textbf{96.1} & \textbf{BATS} & 1/1 \\
    \hline
    \end{tabular}
%}
\end{center}
\caption{Comparison with state-of-the-art binarization/quantization methods on CIFAR-10 across various architecture. Notice that the discovered architecture by our approach significantly outperforms all previous reported results.}
\label{tab:sota-cifar10-all}
\end{table}

\section{Going back to real}

Herein we briefly evaluate the effectiveness of our novel search space and methodology for the case of real-valued networks. To do so, given the proposed search space and temperature regularization mechanism, we performed a network search on CIFAR-10 largely following the procedure described in Section~4, with the following changes: the learning rate for both search and evaluation is set to $0.1$ and the optimizer to SGD with momentum 0.9. As Table~\ref{tab:ev_real_cifar} shows, our method generalizes well to the real-valued case, offering competitive results. This suggests that the operations tailored to binary networks can work well for their real-valued counterparts, too.

\begin{table}[ht!]
\caption{Comparison on the CIFAR-10 dataset for the case of real-valued networks.}
\label{tab:ev_real_cifar}
\centering
%\resizebox{\columnwidth}{!}{%
    \begin{tabular}{lcc}
    \hline
    \textbf{Architecture} & \textbf{Test Err. (\%)} & \textbf{Params (M)}   \\
    \hline
    NASNet-A~\cite{zoph2018learning}                 & 2.65 & 3.3         \\
    AmoebaNet-A~\cite{real2019regularized}           & 3.34 & 3.2        \\
    % Hireachical Evolution~\cite{liu2017hierarchical}          & 3.75 & -     & 15.7 \\
    % PNAS~\cite{liu2018progressive}                            & 3.41 & -     & 3.2\\
    % ENAS + cutout~\cite{pham2018efficient}                    & 2.89 & -     & 4.6\\
    \hline
    DARTS (first order)~\cite{liu2018darts}           & 3.00 & 3.3  \\
    DARTS (second order)~\cite{liu2018darts}          & 2.76 & 3.3   \\
    % SNAS + mild constraint + cutout~\cite{xie2018snas}        & 2.98 & -     & 2.9\\
    % SNAS + moderate constraint + cutout~\cite{xie2018snas}    & 2.85 & -     & 2.8 \\
    % SNAS + aggressive constraint + cutout~\cite{xie2018snas}  & 3.10 & -     & 2.3 \\
    % ProxylessNAS~\cite{cai2018proxylessnas} + cutout   & 2.08 & -     & 5.7 \\
    P-DARTS~\cite{chen2019progressive}                                   & 2.50 & 3.4   \\
    % P-DARTS CIFAR100 + cutout                                   & 2.62 & 15.92 & 3.6  & 0.3 & gradient-based \\
    \hline
    BATS (Ours)                          & 2.70 & 2.5    \\
    BATS (Ours)                          & 2.40 & 3.5    \\
    \hline
    \end{tabular}
%}
%\vspace{-3pt}
\end{table}

\section{Discovered real-valued topologies}

\begin{figure*}[!htpb]
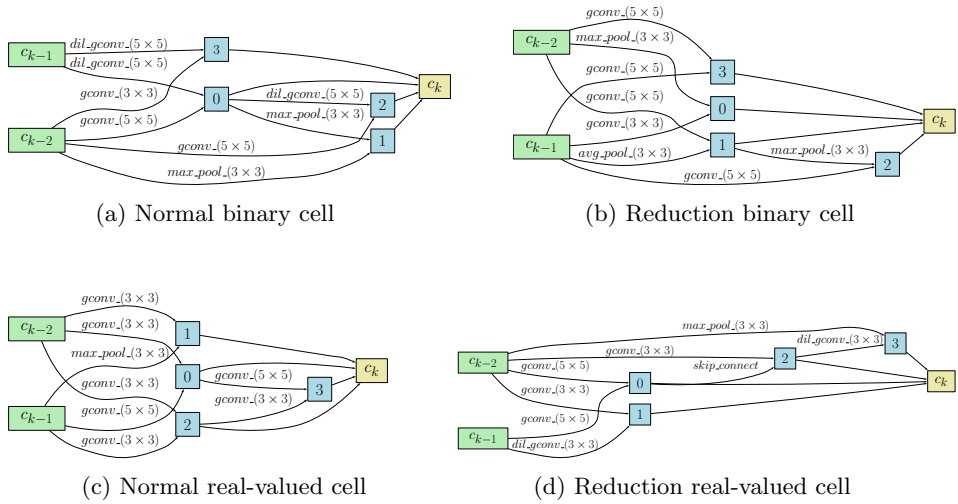

     \centering
     \begin{subfigure}[b]{0.45\textwidth}
         \centering
        % \begin{tikzpicture}[>=latex',line join=bevel]
        %     \useasboundingbox (0,0) rectangle (7.5,3);
        %     \scope[transform canvas={scale=.26}]
        %     \pgfsetlinewidth{1bp}
        %     \input{figures/cells/ours_normal_binary.tex}
        %     \endscope
        % \end{tikzpicture}
        \includegraphics{figures/cells/ours_normal_binary.pdf}
        \caption{Normal binary cell}
        \label{fig:supp_ours_normal_cell}
     \end{subfigure}
     \hfill
     \begin{subfigure}[b]{0.45\textwidth}
        \centering
        % \begin{tikzpicture}[>=latex',line join=bevel]
        %     \useasboundingbox (0,0) rectangle (7.5,3);
        %     \scope[transform canvas={scale=.26}]
        %     \pgfsetlinewidth{1bp}
        %     \input{figures/cells/ours_reduce_binary.tex}
        %     \endscope
        % \end{tikzpicture}
         \includegraphics{figures/cells/ours_reduce_binary.pdf}
        \caption{Reduction binary cell}
        \label{fig:supp_ours_reduction_cell}
     \end{subfigure}
     ~
     \begin{subfigure}[b]{0.47\textwidth}
         \centering
        \begin{tikzpicture}[>=latex',line join=bevel]
            \useasboundingbox (0,0) rectangle (7.5,3);
            \scope[transform canvas={scale=.25}]
            \pgfsetlinewidth{1bp}
            \Huge%
\pgfsetcolor{black}
  % Edge: c_{k-2} -> 0
  \draw [->] (85.336bp,191.43bp) .. controls (137.54bp,187.4bp) and (221.52bp,179.88bp)  .. (234.0bp,172.44bp) .. controls (243.05bp,167.05bp) and (250.35bp,158.3bp)  .. (261.04bp,140.73bp);
  \definecolor{strokecol}{rgb}{0.0,0.0,0.0};
  \pgfsetstrokecolor{strokecol}
  \draw (168.5bp,200.44bp) node {\huge $gconv\_(3\times 3)$};
  % Edge: c_{k-2} -> 1
  \draw [->] (84.541bp,212.54bp) .. controls (90.627bp,214.57bp) and (96.903bp,216.31bp)  .. (103.0bp,217.44bp) .. controls (160.23bp,228.12bp) and (179.39bp,237.64bp)  .. (234.0bp,217.44bp) .. controls (237.78bp,216.05bp) and (241.42bp,214.04bp)  .. (252.81bp,205.46bp);
  \draw (168.5bp,239.44bp) node {\huge $gconv\_(3\times 3)$};
  % Edge: c_{k-2} -> 2
  \draw [->] (49.713bp,176.37bp) .. controls (58.161bp,154.74bp) and (75.456bp,119.21bp)  .. (103.0bp,101.44bp) .. controls (152.44bp,69.555bp) and (180.83bp,107.63bp)  .. (234.0bp,82.443bp) .. controls (237.93bp,80.58bp) and (241.74bp,78.121bp)  .. (253.09bp,68.688bp);
  \draw (168.5bp,112.44bp) node {\huge $gconv\_(3\times 3)$};
  % Edge: c_{k-1} -> 0
  \draw [->] (85.294bp,48.294bp) .. controls (126.15bp,39.875bp) and (188.66bp,33.811bp)  .. (234.0bp,60.443bp) .. controls (246.74bp,67.926bp) and (255.21bp,81.938bp)  .. (264.19bp,104.19bp);
  \draw (168.5bp,71.443bp) node {\huge $gconv\_(5\times 5)$};
  % Edge: c_{k-1} -> 1
  \draw [->] (54.202bp,77.73bp) .. controls (64.556bp,93.64bp) and (81.776bp,115.9bp)  .. (103.0bp,127.44bp) .. controls (154.69bp,155.55bp) and (182.49bp,118.03bp)  .. (234.0bp,146.44bp) .. controls (240.65bp,150.11bp) and (246.6bp,155.57bp)  .. (257.88bp,169.2bp);
  \draw (168.5bp,157.44bp) node {\huge $max\_pool\_(3\times 3)$};
  % Edge: c_{k-1} -> 2
  \draw [->] (61.536bp,41.304bp) .. controls (72.507bp,31.591bp) and (87.382bp,20.591bp)  .. (103.0bp,15.443bp) .. controls (158.3bp,-2.7817bp) and (180.21bp,-6.8376bp)  .. (234.0bp,15.443bp) .. controls (239.36bp,17.663bp) and (244.33bp,21.145bp)  .. (256.03bp,32.436bp);
  \draw (168.5bp,26.443bp) node {\huge $gconv\_(3\times 3)$};
  % Edge: 0 -> 3
  \draw [->] (288.31bp,117.3bp) .. controls (293.91bp,115.82bp) and (300.17bp,114.36bp)  .. (306.0bp,113.44bp) .. controls (353.01bp,106.02bp) and (408.51bp,104.04bp)  .. (450.86bp,103.44bp);
  \draw (369.5bp,124.44bp) node {\huge $gconv\_(5\times 5)$};
  % Edge: 0 -> c_k
  \draw [->] (288.1bp,130.18bp) .. controls (293.69bp,132.33bp) and (300.0bp,134.37bp)  .. (306.0bp,135.44bp) .. controls (379.21bp,148.6bp) and (466.98bp,141.51bp)  .. (523.72bp,134.73bp);
  % Edge: 1 -> c_k
  \draw [->] (288.08bp,184.76bp) .. controls (324.59bp,179.02bp) and (412.94bp,165.12bp)  .. (487.0bp,153.44bp) .. controls (495.44bp,152.11bp) and (497.83bp,152.95bp)  .. (506.0bp,150.44bp) .. controls (508.78bp,149.59bp) and (511.62bp,148.58bp)  .. (523.82bp,143.45bp);
  % Edge: 2 -> 3
  \draw [->] (288.29bp,50.69bp) .. controls (318.74bp,51.689bp) and (383.41bp,56.357bp)  .. (433.0bp,77.443bp) .. controls (436.23bp,78.818bp) and (439.44bp,80.548bp)  .. (450.84bp,88.187bp);
  \draw (369.5bp,88.443bp) node {\huge $gconv\_(3\times 3)$};
  % Edge: 2 -> c_k
  \draw [->] (288.34bp,47.443bp) .. controls (318.34bp,42.867bp) and (381.63bp,35.916bp)  .. (433.0bp,48.443bp) .. controls (469.12bp,57.251bp) and (502.43bp,84.954bp)  .. (530.68bp,113.34bp);
  % Edge: 3 -> c_k
  \draw [->] (487.36bp,109.8bp) .. controls (495.41bp,112.74bp) and (505.24bp,116.34bp)  .. (523.98bp,123.2bp);
  % Node: c_{k-2}
\begin{scope}
  \definecolor{strokecol}{rgb}{0.0,0.0,0.0};
  \pgfsetstrokecolor{strokecol}
  \definecolor{fillcol}{rgb}{0.71,0.93,0.71};
  \pgfsetfillcolor{fillcol}
  \filldraw (85.0bp,212.44bp) -- (0.0bp,212.44bp) -- (0.0bp,176.44bp) -- (85.0bp,176.44bp) -- cycle;
  \draw (42.5bp,194.44bp) node {$c_{k-2}$};
\end{scope}
  % Node: 0
\begin{scope}
  \definecolor{strokecol}{rgb}{0.0,0.0,0.0};
  \pgfsetstrokecolor{strokecol}
  \definecolor{fillcol}{rgb}{0.68,0.85,0.9};
  \pgfsetfillcolor{fillcol}
  \filldraw (288.0bp,140.44bp) -- (252.0bp,140.44bp) -- (252.0bp,104.44bp) -- (288.0bp,104.44bp) -- cycle;
  \draw (270.0bp,122.44bp) node {$0$};
\end{scope}
  % Node: 1
\begin{scope}
  \definecolor{strokecol}{rgb}{0.0,0.0,0.0};
  \pgfsetstrokecolor{strokecol}
  \definecolor{fillcol}{rgb}{0.68,0.85,0.9};
  \pgfsetfillcolor{fillcol}
  \filldraw (288.0bp,205.44bp) -- (252.0bp,205.44bp) -- (252.0bp,169.44bp) -- (288.0bp,169.44bp) -- cycle;
  \draw (270.0bp,187.44bp) node {$1$};
\end{scope}
  % Node: 2
\begin{scope}
  \definecolor{strokecol}{rgb}{0.0,0.0,0.0};
  \pgfsetstrokecolor{strokecol}
  \definecolor{fillcol}{rgb}{0.68,0.85,0.9};
  \pgfsetfillcolor{fillcol}
  \filldraw (288.0bp,68.44bp) -- (252.0bp,68.44bp) -- (252.0bp,32.44bp) -- (288.0bp,32.44bp) -- cycle;
  \draw (270.0bp,50.443bp) node {$2$};
\end{scope}
  % Node: c_{k-1}
\begin{scope}
  \definecolor{strokecol}{rgb}{0.0,0.0,0.0};
  \pgfsetstrokecolor{strokecol}
  \definecolor{fillcol}{rgb}{0.71,0.93,0.71};
  \pgfsetfillcolor{fillcol}
  \filldraw (85.0bp,77.44bp) -- (0.0bp,77.44bp) -- (0.0bp,41.44bp) -- (85.0bp,41.44bp) -- cycle;
  \draw (42.5bp,59.443bp) node {$c_{k-1}$};
\end{scope}
  % Node: 3
\begin{scope}
  \definecolor{strokecol}{rgb}{0.0,0.0,0.0};
  \pgfsetstrokecolor{strokecol}
  \definecolor{fillcol}{rgb}{0.68,0.85,0.9};
  \pgfsetfillcolor{fillcol}
  \filldraw (487.0bp,121.44bp) -- (451.0bp,121.44bp) -- (451.0bp,85.44bp) -- (487.0bp,85.44bp) -- cycle;
  \draw (469.0bp,103.44bp) node {$3$};
\end{scope}
  % Node: c_k
\begin{scope}
  \definecolor{strokecol}{rgb}{0.0,0.0,0.0};
  \pgfsetstrokecolor{strokecol}
  \definecolor{fillcol}{rgb}{0.93,0.91,0.67};
  \pgfsetfillcolor{fillcol}
  \filldraw (571.0bp,149.44bp) -- (524.0bp,149.44bp) -- (524.0bp,113.44bp) -- (571.0bp,113.44bp) -- cycle;
  \draw (547.5bp,131.44bp) node {$c_k$};
\end{scope}
            \endscope
        \end{tikzpicture}
        \caption{Normal real-valued cell}
        \label{fig:supp_ours_normal_cell_real}
     \end{subfigure}
     \hfill
     \begin{subfigure}[b]{0.51\textwidth}
        \centering
        \begin{tikzpicture}[>=latex',line join=bevel]
            \useasboundingbox (0,0) rectangle (7.5,3);
            \scope[transform canvas={scale=.22}]
            \pgfsetlinewidth{1bp}
            \Huge%
\pgfsetcolor{black}
  % Edge: c_{k-2} -> 0
  \draw [->] (85.194bp,150.16bp) .. controls (91.129bp,148.62bp) and (97.189bp,147.23bp)  .. (103.0bp,146.16bp) .. controls (166.87bp,134.37bp) and (243.26bp,130.26bp)  .. (293.41bp,128.53bp);
  \definecolor{strokecol}{rgb}{0.0,0.0,0.0};
  \pgfsetstrokecolor{strokecol}
  \draw (166.5bp,157.16bp) node {\huge $gconv\_(5\times 5)$};
  % Edge: c_{k-2} -> 1
  \draw [->] (56.789bp,145.14bp) .. controls (67.662bp,131.82bp) and (84.261bp,114.45bp)  .. (103.0bp,105.16bp) .. controls (134.17bp,89.695bp) and (234.38bp,80.046bp)  .. (293.38bp,75.358bp);
  \draw (166.5bp,116.16bp) node {\huge $gconv\_(3\times 3)$};
  % Edge: c_{k-2} -> 2
  \draw [->] (85.085bp,167.07bp) .. controls (91.091bp,167.52bp) and (97.201bp,167.9bp)  .. (103.0bp,168.16bp) .. controls (265.16bp,175.31bp) and (460.07bp,172.29bp)  .. (541.59bp,170.58bp);
  \draw (311.5bp,183.16bp) node {\huge $gconv\_(3\times 3)$};
  % Edge: c_{k-2} -> 3
  \draw [->] (85.108bp,178.43bp) .. controls (91.056bp,180.24bp) and (97.143bp,181.89bp)  .. (103.0bp,183.16bp) .. controls (166.31bp,196.9bp) and (183.41bp,193.06bp)  .. (248.0bp,198.16bp) .. controls (402.42bp,210.35bp) and (441.16bp,212.93bp)  .. (596.0bp,217.16bp) .. controls (647.98bp,218.57bp) and (662.89bp,231.04bp)  .. (713.0bp,217.16bp) .. controls (715.98bp,216.33bp) and (718.96bp,215.18bp)  .. (730.69bp,209.05bp);
  \draw (458.5bp,225.16bp) node {\huge $max\_pool\_(3\times 3)$};
  % Edge: c_{k-1} -> 0
  \draw [->] (85.075bp,33.202bp) .. controls (133.02bp,34.074bp) and (208.07bp,38.218bp)  .. (230.0bp,55.156bp) .. controls (247.38bp,68.575bp) and (232.75bp,85.365bp)  .. (248.0bp,101.16bp) .. controls (257.56bp,111.05bp) and (271.35bp,117.53bp)  .. (293.28bp,124.59bp);
  \draw (166.5bp,66.156bp) node {\huge $gconv\_(5\times 5)$};
  % Edge: c_{k-1} -> 1
  \draw [->] (85.24bp,15.637bp) .. controls (91.114bp,13.789bp) and (97.146bp,12.204bp)  .. (103.0bp,11.156bp) .. controls (158.56bp,1.2135bp) and (176.64bp,-7.2419bp)  .. (230.0bp,11.156bp) .. controls (251.86bp,18.693bp) and (272.32bp,34.877bp)  .. (294.36bp,56.066bp);
  \draw (166.5bp,22.156bp) node {\huge$dil\_gconv\_(3\times 3)$};
  % Edge: 0 -> 2
  \draw [->] (329.57bp,126.73bp) .. controls (366.34bp,124.27bp) and (455.28bp,121.62bp)  .. (524.0bp,146.16bp) .. controls (527.12bp,147.27bp) and (530.23bp,148.74bp)  .. (541.81bp,155.78bp);
  \draw (458.5bp,157.16bp) node {\huge$skip\_connect$};
  % Edge: 0 -> c_k
  \draw [->] (329.77bp,127.6bp) .. controls (346.07bp,127.11bp) and (371.19bp,126.43bp)  .. (393.0bp,126.16bp) .. controls (542.61bp,124.3bp) and (721.27bp,128.34bp)  .. (803.94bp,130.52bp);
  % Edge: 1 -> c_k
  \draw [->] (329.73bp,75.762bp) .. controls (391.56bp,81.578bp) and (608.03bp,102.4bp)  .. (786.0bp,125.16bp) .. controls (788.53bp,125.48bp) and (791.15bp,125.83bp)  .. (803.82bp,127.63bp);
  % Edge: 2 -> 3
  \draw [->] (578.26bp,172.56bp) .. controls (610.82bp,177.08bp) and (681.56bp,186.92bp)  .. (730.78bp,193.76bp);
  \draw (654.5bp,202.16bp) node {\huge$dil\_gconv\_(3\times 3)$};
  % Edge: 2 -> c_k
  \draw [->] (578.01bp,167.66bp) .. controls (621.43bp,161.28bp) and (736.94bp,144.31bp)  .. (803.78bp,134.49bp);
  % Edge: 3 -> c_k
  \draw [->] (767.36bp,181.41bp) .. controls (776.19bp,173.9bp) and (787.17bp,164.57bp)  .. (805.11bp,149.33bp);
  % Node: c_{k-2}
\begin{scope}
  \definecolor{strokecol}{rgb}{0.0,0.0,0.0};
  \pgfsetstrokecolor{strokecol}
  \definecolor{fillcol}{rgb}{0.71,0.93,0.71};
  \pgfsetfillcolor{fillcol}
  \filldraw (85.0bp,181.16bp) -- (0.0bp,181.16bp) -- (0.0bp,145.16bp) -- (85.0bp,145.16bp) -- cycle;
  \draw (42.5bp,163.16bp) node {$c_{k-2}$};
\end{scope}
  % Node: 0
\begin{scope}
  \definecolor{strokecol}{rgb}{0.0,0.0,0.0};
  \pgfsetstrokecolor{strokecol}
  \definecolor{fillcol}{rgb}{0.68,0.85,0.9};
  \pgfsetfillcolor{fillcol}
  \filldraw (329.5bp,146.16bp) -- (293.5bp,146.16bp) -- (293.5bp,110.16bp) -- (329.5bp,110.16bp) -- cycle;
  \draw (311.5bp,128.16bp) node {$0$};
\end{scope}
  % Node: 1
\begin{scope}
  \definecolor{strokecol}{rgb}{0.0,0.0,0.0};
  \pgfsetstrokecolor{strokecol}
  \definecolor{fillcol}{rgb}{0.68,0.85,0.9};
  \pgfsetfillcolor{fillcol}
  \filldraw (329.5bp,92.16bp) -- (293.5bp,92.16bp) -- (293.5bp,56.16bp) -- (329.5bp,56.16bp) -- cycle;
  \draw (311.5bp,74.156bp) node {$1$};
\end{scope}
  % Node: 2
\begin{scope}
  \definecolor{strokecol}{rgb}{0.0,0.0,0.0};
  \pgfsetstrokecolor{strokecol}
  \definecolor{fillcol}{rgb}{0.68,0.85,0.9};
  \pgfsetfillcolor{fillcol}
  \filldraw (578.0bp,188.16bp) -- (542.0bp,188.16bp) -- (542.0bp,152.16bp) -- (578.0bp,152.16bp) -- cycle;
  \draw (560.0bp,170.16bp) node {$2$};
\end{scope}
  % Node: 3
\begin{scope}
  \definecolor{strokecol}{rgb}{0.0,0.0,0.0};
  \pgfsetstrokecolor{strokecol}
  \definecolor{fillcol}{rgb}{0.68,0.85,0.9};
  \pgfsetfillcolor{fillcol}
  \filldraw (767.0bp,214.16bp) -- (731.0bp,214.16bp) -- (731.0bp,178.16bp) -- (767.0bp,178.16bp) -- cycle;
  \draw (749.0bp,196.16bp) node {$3$};
\end{scope}
  % Node: c_{k-1}
\begin{scope}
  \definecolor{strokecol}{rgb}{0.0,0.0,0.0};
  \pgfsetstrokecolor{strokecol}
  \definecolor{fillcol}{rgb}{0.71,0.93,0.71};
  \pgfsetfillcolor{fillcol}
  \filldraw (85.0bp,51.16bp) -- (0.0bp,51.16bp) -- (0.0bp,15.16bp) -- (85.0bp,15.16bp) -- cycle;
  \draw (42.5bp,33.156bp) node {$c_{k-1}$};
\end{scope}
  % Node: c_k
\begin{scope}
  \definecolor{strokecol}{rgb}{0.0,0.0,0.0};
  \pgfsetstrokecolor{strokecol}
  \definecolor{fillcol}{rgb}{0.93,0.91,0.67};
  \pgfsetfillcolor{fillcol}
  \filldraw (851.0bp,149.16bp) -- (804.0bp,149.16bp) -- (804.0bp,113.16bp) -- (851.0bp,113.16bp) -- cycle;
  \draw (827.5bp,131.16bp) node {$c_k$};
\end{scope}
            \endscope
        \end{tikzpicture}
        \caption{Reduction real-valued cell}
        \label{fig:supp_ours_reduction_cell_real}
     \end{subfigure}
     \caption{Normal and reduction cells discovered by our proposed method using the introduced search space for the binary case (first row) and real-valued case (second row). Notice that the binary cells tend to be shallower and to contain convolutional operations with larger kernels (i.e. $5\times 5$) when compared with the real-valued ones.}
     \label{fig:all-cells}
\end{figure*}

While the proposed search space and method is mainly geared towards binary networks, we also tested its generalizability on the real-valued domain. Fig.~\ref{fig:supp_ours_normal_cell_real} and~\ref{fig:supp_ours_reduction_cell_real} depict an example of cells found by our approach when using real valued networks. Notice that as opposed to the binary ones (Fig.~\ref{fig:supp_ours_normal_cell} and~\ref{fig:supp_ours_reduction_cell}), the real valued ones tend to be deeper and use operations with smaller convolutional kernels.

 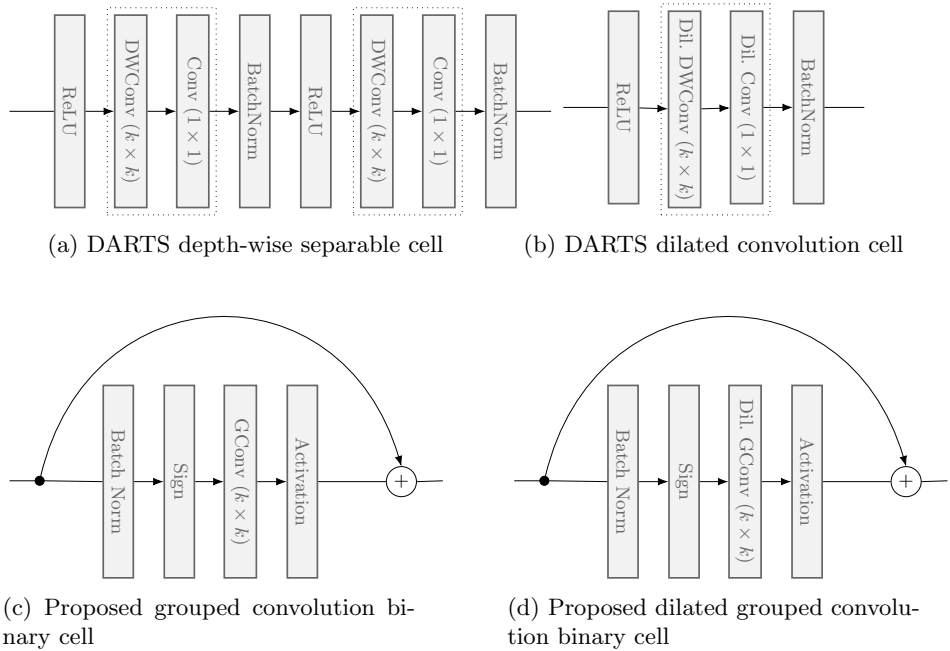
\begin{figure*}[!htpb]
     \centering
     \begin{subfigure}[b]{0.53\textwidth}
         \centering
        \scalebox{0.8}{\def\X{1}%
\def\Y{2}%
\def\WIDTH{0.5}%
\def\HEIGHT{3.2}%
\def\D{0.75}

\begin{tikzpicture}
\node[draw, color=black!60, fill=black!5, thick, rotate=270, minimum height=\WIDTH cm, minimum width=\HEIGHT cm] at (\X, \Y) (r1) {\small ReLU};
\node[draw, color=black!60, fill=black!5, thick, rotate=270, minimum height=\WIDTH cm, minimum width=\HEIGHT cm, right=\D cm of r1.north west]  (c1) {\small DWConv ($k\times k$)};
\node[draw, color=black!60, fill=black!5, thick, rotate=270, minimum height=\WIDTH cm, minimum width=\HEIGHT cm, right=\D cm of c1.north west] (c2) {\small Conv ($1\times 1$)};
\node[draw, color=black!60, fill=black!5, thick, rotate=270, minimum height=\WIDTH cm, minimum width=\HEIGHT cm, right=\D cm of c2.north west] (b1) {\small BatchNorm};

\node[draw, color=black!60, fill=black!5, thick, rotate=270, minimum height=\WIDTH cm, minimum width=\HEIGHT cm, right=\D cm of b1.north west] (r2) {\small ReLU};
\node[draw, color=black!60, fill=black!5, thick, rotate=270, minimum height=\WIDTH cm, minimum width=\HEIGHT cm, right=\D cm of r2.north west] (c3) {\small DWConv ($k\times k$)};
\node[draw, color=black!60, fill=black!5, thick, rotate=270, minimum height=\WIDTH cm, minimum width=\HEIGHT cm, right=\D cm of c3.north west] (c4) {\small Conv ($1\times 1$)};
\node[draw, color=black!60, fill=black!5, thick, rotate=270, minimum height=\WIDTH cm, minimum width=\HEIGHT cm, right=\D cm of c4.north west] (b2) {\small BatchNorm};

\draw [-] (0,2) edge (r1.south);
\draw [-Latex] (r1) edge (c1) (c1) edge (c2) (c2) edge (b1) (b1) edge (r2) (r2) edge (c3) (c3) edge (c4) (c4) edge (b2);
\draw [-] (b2.north) edge (9,2);

\node[draw,dotted,fit=(c1) (c2)] {};
\node[draw,dotted,fit=(c3) (c4)] {};
\end{tikzpicture}}
        \caption{DARTS depth-wise separable cell}
        \label{fig:real_sep_conv}
     \end{subfigure}
     \hfill
     \begin{subfigure}[b]{0.45\textwidth}
        \centering
        \scalebox{0.8}{\def\X{1}%
\def\Y{2}%
\def\WIDTH{0.5}%
\def\HEIGHT{3.2}%
\def\D{0.75}

\begin{tikzpicture}
\node[draw, color=black!60, fill=black!5, thick, rotate=270, minimum height=\WIDTH cm, minimum width=\HEIGHT cm] at (\X, \Y) (r1) {\small ReLU};
\node[draw, color=black!60, fill=black!5, thick, rotate=270, minimum height=\WIDTH cm, minimum width=\HEIGHT cm, right=\D cm of r1.north west]  (c1) {\small Dil. DWConv ($k\times k$)};
\node[draw, color=black!60, fill=black!5, thick, rotate=270, minimum height=\WIDTH cm, minimum width=\HEIGHT cm, right=\D cm of c1.north west] (c2) {\small Dil. Conv ($1\times 1$)};
\node[draw, color=black!60, fill=black!5, thick, rotate=270, minimum height=\WIDTH cm, minimum width=\HEIGHT cm, right=\D cm of c2.north west] (b1) {\small BatchNorm};

\draw [-] (0,2) edge (r1.south);
\draw [-Latex] (r1) edge (c1) (c1) edge (c2) (c2) edge (b1);
\draw [-] (b1.north) edge (5,2);

\node[draw,dotted,fit=(c1) (c2)] {};
\end{tikzpicture}}
        \caption{DARTS dilated convolution cell}
        \label{fig:real_dil_conv}
     \end{subfigure}
     ~
     \begin{subfigure}[b]{0.45\textwidth}
         \centering
        \scalebox{0.8}{\def\X{1}%
\def\Y{2}%
\def\WIDTH{0.5}%
\def\HEIGHT{3.2}%
\def\D{0.75}

\begin{tikzpicture}

\node[circle, draw=black, fill=black, inner sep=0cm, minimum size=0.15cm] at (\X, \Y) (in)  {};

\node[draw, color=black!60, fill=black!5, thick, rotate=270, minimum height=\WIDTH cm, minimum width=\HEIGHT cm, above right= 1.60 cm  and \D*1.5 cm of in.west] (bn) {\small Batch Norm};
\node[draw, color=black!60, fill=black!5, thick, rotate=270, minimum height=\WIDTH cm, minimum width=\HEIGHT cm, right=\D cm of bn.north west]  (sign) {\small Sign};
\node[draw, color=black!60, fill=black!5, thick, rotate=270, minimum height=\WIDTH cm, minimum width=\HEIGHT cm, right=\D cm of sign.north west] (conv) {\small GConv ($k\times k$)};
\node[draw, color=black!60, fill=black!5, thick, rotate=270, minimum height=\WIDTH cm, minimum width=\HEIGHT cm, right=\D cm of conv.north west] (act) {\small Activation};

\node[circle, draw=black, fill=white, inner sep=0cm, minimum size=0.5cm, right=\D*1.5 cm of act.north] (out)  {$+$};

\draw [-] (0.5,2) edge (in.east);
\draw [-] (in.east) edge (bn.south);
\draw [-Latex] (bn) edge (sign) (sign) edge (conv) (conv) edge (act);
\draw [-] (act.north) edge (out.west);
\draw [-] (out.east) edge (7.7,2);
%\draw[-Latex] let \p{sign}=(sign.west), \p{conv}=(conv.west) in (in) to  [black,bend left] (\x{sign}/2+\x{conv}/2, \y{sign}+1cm) to [black, bend left] (out.north); 
\draw[-Latex] let \p{sign}=(sign.west), \p{conv}=(conv.west) in (in) to  [black,bend left,out=75,in=105,looseness=1.5]  (out.north); 
%\draw[-Latex] let \p{sign}=(sign.west), \p{conv}=(conv.west) in (\x{sign}/2+\x{conv}/2, \y{sign}+1cm) edge  [black,bend left] (out.north) ;

\end{tikzpicture}}
        \caption{Proposed grouped convolution binary cell}
        \label{fig:bin_sep_conv}
     \end{subfigure}
     \hfill
     \begin{subfigure}[b]{0.45\textwidth}
        \centering
        \scalebox{0.8}{\def\X{1}%
\def\Y{2}%
\def\WIDTH{0.5}%
\def\HEIGHT{3.2}%
\def\D{0.75}

\begin{tikzpicture}

\node[circle, draw=black, fill=black, inner sep=0cm, minimum size=0.15cm] at (\X, \Y) (in)  {};

\node[draw, color=black!60, fill=black!5, thick, rotate=270, minimum height=\WIDTH cm, minimum width=\HEIGHT cm, above right= 1.60 cm  and \D*1.5 cm of in.west] (bn) {\small Batch Norm};
\node[draw, color=black!60, fill=black!5, thick, rotate=270, minimum height=\WIDTH cm, minimum width=\HEIGHT cm, right=\D cm of bn.north west]  (sign) {\small Sign};
\node[draw, color=black!60, fill=black!5, thick, rotate=270, minimum height=\WIDTH cm, minimum width=\HEIGHT cm, right=\D cm of sign.north west] (conv) {\small Dil. GConv ($k\times k$)};
\node[draw, color=black!60, fill=black!5, thick, rotate=270, minimum height=\WIDTH cm, minimum width=\HEIGHT cm, right=\D cm of conv.north west] (act) {\small Activation};

\node[circle, draw=black, fill=white, inner sep=0cm, minimum size=0.5cm, right=\D*1.5 cm of act.north] (out)  {$+$};

\draw [-] (0.5,2) edge (in.east);
\draw [-] (in.east) edge (bn.south);
\draw [-Latex] (bn) edge (sign) (sign) edge (conv) (conv) edge (act);
\draw [-] (act.north) edge (out.west);
\draw [-] (out.east) edge (7.7,2);
%\draw[-Latex] let \p{sign}=(sign.west), \p{conv}=(conv.west) in (in) to  [black,bend left] (\x{sign}/2+\x{conv}/2, \y{sign}+1cm) to [black, bend left] (out.north); 
\draw[-Latex] let \p{sign}=(sign.west), \p{conv}=(conv.west) in (in) to  [black,bend left,out=75,in=105,looseness=1.5]  (out.north); 
%\draw[-Latex] let \p{sign}=(sign.west), \p{conv}=(conv.west) in (\x{sign}/2+\x{conv}/2, \y{sign}+1cm) edge  [black,bend left] (out.north) ;

\end{tikzpicture}}
        \caption{Proposed dilated grouped convolution binary cell}
        \label{fig:bin_dil_conv}
     \end{subfigure}
     \caption{Comparison between the convolutional operations used in the DARTS search space (\ref{fig:real_sep_conv} and~\ref{fig:real_dil_conv}) and the proposed ones (\ref{fig:bin_sep_conv} and~\ref{fig:bin_dil_conv}). $k \times k$ denotes the kernel size, \protect\tikz  \protect\node[circle, draw=black, fill=white, inner sep=0cm, minimum size=0.35cm] (out)  {$+$}; is the element-wise summation operation while each rectangle represents a given operation defined by the inner text.}
     \label{fig:real_vs_bin_ops}
\end{figure*}

\end{document}